%% file: main.tex
\documentclass{article}
\PassOptionsToPackage{dvipsnames}{xcolor}
\usepackage{microtype}
\usepackage{graphicx}
\usepackage{subfigure}
\usepackage{booktabs} % for professional tables

\usepackage{hyperref}

\usepackage[accepted]{icml2025}

\usepackage{amsmath}
\usepackage{amssymb}
\usepackage{mathtools}
\usepackage{amsthm}

\usepackage[capitalize,noabbrev]{cleveref}

\theoremstyle{plain}
\newtheorem{theorem}{Theorem}[section]

\theoremstyle{definition}

\theoremstyle{remark}

\usepackage{colortbl}

\usepackage[textsize=tiny]{todonotes}

\icmltitlerunning{Info-Coevolution: An Efficient Framework for Data Model Coevolution}

\begin{document}

\twocolumn[
\icmltitle{Info-Coevolution: An Efficient Framework for Data Model Coevolution}



\icmlsetsymbol{core contribution}{$\S$}
\icmlsetsymbol{corresponding}{$\dagger$}
\icmlsetsymbol{project_lead}{$\diamondsuit$}

\begin{icmlauthorlist}
\icmlauthor{Ziheng Qin}{NUS,TikTok}
\icmlauthor{Hailun Xu}{TikTok,core contribution}
\icmlauthor{Wei Chee Yew}{TikTok}
\icmlauthor{Qi Jia}{NUS}
\icmlauthor{Yang Luo}{NUS}
\icmlauthor{Kanchan Sarkar}{TikTok,project_lead}
\icmlauthor{Danhui Guan}{TikTok,project_lead}
\icmlauthor{Kai Wang}{NUS,core contribution}
\icmlauthor{Yang You}{NUS,corresponding}
\end{icmlauthorlist}

\icmlaffiliation{NUS}{National University of Singapore, School of Computing, 11 Research Link, Singapore}
\icmlaffiliation{TikTok}{ByteDance Ltd, Singapore}

\icmlcorrespondingauthor{Ziheng Qin}{zihengq@comp.nus.edu.sg}
\icmlcorrespondingauthor{Kanchan Sarkar}{kanchan.sarkar@bytedance.com}
\icmlcorrespondingauthor{Danhui Guan}{guandanhui@bytedance.com}
\icmlcorrespondingauthor{Yang You}{youy@comp.nus.edu.sg}

\icmlkeywords{Machine Learning, Active Learning, Large Scale, Annotation Efficiency}

\vskip 0.3in
]



\printAffiliationsAndNotice{\icmlFootNotes}  % leave blank if no need to mention equal contribution


% 梳理一下，我们现在主要贡献有：1.提供了一个理论框架和对应的在线（主动学习）方法，基于模型和样本的共同预测，关于样本到底有没有用的一个定性值，能够efficient地扩展。在imagenet训练上能节省30%标注（不需要提前知道label），能实现数据和模型的高效共同提升 
% 2. 分析了半监督方法的一些问题 X。 
% 2. 简单说下半监督方法的问题，能否无损依赖于数据集特性，也无法确定需要多少数据
% 3. 对annotation cost和efficiency进行了分析以及。   实践性的研究和改进

% 1. 拓展了理论 2. 应用理论提出了方法，更精准地评估了定性的gain，实验证明xxxx，开源。。。 3. 拓展数据的利用

% 几个核心输出观点排列下顺序：拓展的理论部分，方法，有效性；cost and performance scaling pareto, 以及对应的数据量上限; 增长数据和能用来训练的数据；

\begin{abstract}
Machine learning relies heavily on data, yet the continuous growth of real-world data poses challenges for efficient dataset construction and training. A fundamental yet unsolved question is: given our current model and data, does a new data (sample/batch) need annotation/learning?
Conventional approaches retain all available data, leading to non-optimal data and training efficiency. Active learning aims to reduce data redundancy by selecting a subset of samples to annotate, while it increases pipeline complexity and introduces bias.
In this work, we propose Info-Coevolution, a novel framework that efficiently enables models and data to coevolve through online selective annotation with no bias. Leveraging task-specific models (and open-source models), it selectively annotates and integrates online and web data to improve datasets efficiently. For real-world datasets like ImageNet-1K, Info-Coevolution reduces annotation and training costs by 32\% without performance loss. It is able to automatically give the saving ratio without tuning the ratio. It can further reduce the annotation ratio to 50\% with semi-supervised learning. We also explore retrieval-based dataset enhancement using unlabeled open-source data. Code is available at \href{https://github.com/NUS-HPC-AI-Lab/Info-Coevolution/}{https://github.com/NUS-HPC-AI-Lab/Info-Coevolution/}.
% With the new paradigm, we are able to construct an ImageNet-scale dataset with 1 8-GPU server in 1 week by 1 person.

\end{abstract}

\newcommand{\green}[1]{$_{\color{Green}\uparrow #1}$}

\input{sections/1-introduction}
\input{sections/2-background}
\input{sections/3-method}
\input{sections/4-analysis}
\input{sections/5-conclusion}

\textbf{Acknowledgement}
This work is supported by Bytedance, and was done when Ziheng Qin intern at Bytedance. Ziheng Qin, Hailun Xu (experiments) and Kai Wang (writing) are core contributors. Kanchan Sarkar and Danhui Guan are project leads. Kai Wang is supported by the National Research Foundation, Singapore under its AI Singapore Programme (AISG Award No: AISG2-PhD-2021-08-008). Yang You's research group is being sponsored by NUS startup grant (Presidential Young Professorship), Singapore MOE Tier-1 grant, ByteDance grant, ARCTIC grant, SMI grant (WBS number: A-8001104-00-00),  Alibaba grant, and Google grant for TPU usage. We thank Dawei Du for valuable discussions and feedback.

\newpage
\section*{Impact Statement}

This paper presents research aimed at advancing the field of efficient training in the context of continuously growing datasets. We conducted experiments using publicly available, widely-used datasets in machine learning, without incorporating any new data. As such, our work poses no risk of harmful societal consequences. Instead, it contributes to society by enabling the development of better neural models at a lower cost.

% Authors are \textbf{required} to include a statement of the potential 
% broader impact of their work, including its ethical aspects and future 
% societal consequences. This statement should be in an unnumbered 
% section at the end of the paper (co-located with Acknowledgements -- 
% the two may appear in either order, but both must be before References), 
% and does not count toward the paper page limit. In many cases, where 
% the ethical impacts and expected societal implications are those that 
% are well established when advancing the field of Machine Learning, 
% substantial discussion is not required, and a simple statement such 
% as the following will suffice:

% ``This paper presents work whose goal is to advance the field of 
% Machine Learning. There are many potential societal consequences 
% of our work, none which we feel must be specifically highlighted here.''

% The above statement can be used verbatim in such cases, but we 
% encourage authors to think about whether there is content which does 
% warrant further discussion, as this statement will be apparent if the 
% paper is later flagged for ethics review.


\bibliography{main}
\bibliographystyle{icml2025}

%%%%%%%%%%%%%%%%%%%%%%%%%%%%%%%%%%%%%%%%%%%%%%%%%%%%%%%%%%%%%%%%%%%%%%%%%%%%%%%
%%%%%%%%%%%%%%%%%%%%%%%%%%%%%%%%%%%%%%%%%%%%%%%%%%%%%%%%%%%%%%%%%%%%%%%%%%%%%%%
% APPENDIX
%%%%%%%%%%%%%%%%%%%%%%%%%%%%%%%%%%%%%%%%%%%%%%%%%%%%%%%%%%%%%%%%%%%%%%%%%%%%%%%
%%%%%%%%%%%%%%%%%%%%%%%%%%%%%%%%%%%%%%%%%%%%%%%%%%%%%%%%%%%%%%%%%%%%%%%%%%%%%%%
\newpage
\appendix
\onecolumn

\section{Proof}
We show our proof of Theorem\ref{lemma_sim} here as follows:
For the model $M=g\circ f$, assume $g$ is $L_g$-Lipschitz, i.e. for all $z_1$,$z_2$ in the feature space, we have
\begin{equation}
    ||g(z_1)-g(z_2)||\leq L_g||z_1-z_2||
\end{equation}
Then,
\begin{equation}
    ||g(z_1)-g(z_2)||\leq L_g||z_1-z_2||\leq L_g\epsilon
\end{equation}

% If g is a single linear layer $g(z)=w^Tz+b$, then $L_g$ is bounded by the max norm of w $||W||_m$. 
Softmax is known to be 1-Lipschitz which does not further change the bound of logits.

When using cosine distance, we know that
\begin{equation}
    cosine\_dis(z_1,z_2)=1-\frac{<z_1,z_2>}{||z_1||||z_2||}
\end{equation}

If features $||z_1||$,$||z_2||$ are normalized (as in ViT networks), then 
\begin{equation}
    ||z_1-z_2||=\sqrt{2*cosine\_dist(z_1,z_2)}
\end{equation}
And
\begin{equation}
    cosine\_dist(z_1,z_2)<\epsilon \Rightarrow ||g(z_1)-g(z_2)||<=L_g\sqrt{2\epsilon}
\end{equation}

\section{Additional Experiment Details}
We follow the experimental settings in \href{https://github.com/amazon-science/semi-vit.git}{Semi-ViT}. Our results are trained on a single node of 8 NVIDIA A100-SXM4-80G. For supervised finetuning, we train the model with batchsize 512, learning rate 0.001 for 50 epochs with all augmentations same as in Semi-ViT.

\section{Semi-supervised Data Selection}
For selecting the semi-supervised data, we further add another term based on each sample's average distance to nearest k neighbours in high-confidence samples and already selected samples.
\begin{equation}
    Semi-Gain(z) = \sum_{x \in KNN, x\in{Selected}} cosine\_dist(z,x)
\end{equation}
This term is to encourage a more uniform distribution of unlabeled samples in the space between the labeled samples, so that the semi-supervised training can progressively learn the pseudo target.

% \section{Code}


% You can have as much text here as you want. The main body must be at most $8$ pages long.
% For the final version, one more page can be added.
% If you want, you can use an appendix like this one.  

% The $\mathtt{\backslash onecolumn}$ command above can be kept in place if you prefer a one-column appendix, or can be removed if you prefer a two-column appendix.  Apart from this possible change, the style (font size, spacing, margins, page numbering, etc.) should be kept the same as the main body.
%%%%%%%%%%%%%%%%%%%%%%%%%%%%%%%%%%%%%%%%%%%%%%%%%%%%%%%%%%%%%%%%%%%%%%%%%%%%%%%
%%%%%%%%%%%%%%%%%%%%%%%%%%%%%%%%%%%%%%%%%%%%%%%%%%%%%%%%%%%%%%%%%%%%%%%%%%%%%%%

\end{document}

%% file: sections/1-introduction.tex
\section{Introduction}
% 1. 研究的必要性：机器学习是data-driven的，也有很多实际场景应用。关于数据构造的方面，目前（学界/业界）的主流范式是要么收集+标注加全监督，要么data from all web sources + 自动清洗/无监督+下游监督微调训练。但是两种开销都是巨大的，只有少数的业界/学界机构能负担，大规模数据集构造不是一个足够民主和去中心化的东西。
% 2. 我们想解决的问题，和现在有哪些方案，以及其不足：因此，我们希望能提供一个切实可行的低成本数据集构造方案，能有效在工业/学术的下游场景，进行基于在线数据和网络数据的数据集构建/增强。目前比较相关的研究有（比如imagenet-R...robot领域的...）但是这些项目都开销巨大。半监督标注开销更低但是不能继续scale...
% 3. 因此，我们提出了Data Model Coevolution, 。。。。。。 核心是model pred entropy + distance entropy, 原理是universal approximation theorem

Machine learning is inherently data-driven and has numerous practical applications. Currently, the dominant paradigms for large-scale dataset construction fall into two categories: (1) collecting and annotating data for fully supervised training, or (2) sourcing data from the web, followed by automated cleaning and weakly-supervised/unsupervised training, and subsequently fine-tuning models on downstream tasks with supervised data. However, both approaches incur significant costs for both annotation and training, making them viable only for a few well-resourced institutions. As a result, the construction of large-scale datasets remains an insufficiently democratized and decentralized process.

%分为两段： 现在有哪些比较有影响力的dataset (MNIST, CIFAR, ImageNet, COCO, CC3M CC12M, LAION, SAM-1B robotic ones...)，supervised程度和size，以及构建开销. 人工标注的，weak标注的，无监督的（但依旧有质量差异）。可能需要画张图。一般来说全监督数据的构建开销是超过训练开销的。单纯学界能贡献的数据集（最大的是imagenet？Laion？），大部分是alpaca这个规模。
%第一段，imagenet
Before 2010, popular datasets such as MNIST \cite{mnist6296535} and CIFAR-10/100 \cite{cifar10,cifar100} each contained 60,000 hand-annotated low-resolution images, reflecting the modest scale of early machine learning datasets. The lack of data limited the development speed of machine learning at that time.
In 2010, ImageNet-1K \cite{imagenet5206848} was publicly released with more than 1M human annotated samples. It was the first large-scale supervised dataset, leading to the development of many modern deep-learning algorithms and architectures, boosting the development of AI research today. An estimated cost of building ImageNet1-k is at least 0.4 million USD for labeling the total 1.28m images, and 4 million for 14m ImageNet-12k, excluding other costs. 
With cloud GPU today, training a deep learning neural network on ImageNet-1K with A100 GPU takes about \$240 to \$1400, while labeling an ImageNet-scale (million-scale) data still takes more than 100 times this cost, ranging from \$24,000 to more than \$1m depending on annotators' degree of proficiency. Constructing supervised data is usually more expensive compared to training with them.

%第二段，ImageNet以后其他的数据集，提一下coco，后面的主流有弱监督/无监督的，还有后来各种半监督和模拟数据. weak-supervised data scaling 是一种方式；SAM-1B 是另一种 (human in the loop model-data coevolution)，但是not-affordable by most.
Besides the supervised datasets, other types of datasets emerge later with corresponding training schemes. Bert \cite{devlin2019bertpretrainingdeepbidirectional} and GPT \cite{brown2020languagemodelsfewshotlearners} use large-scale unsupervised text corpus, such as BooksCorpus~\cite{zhu2015aligning} and Common Crawl~\cite{raffel2020exploring}, to train with masked language modeling/next-token-prediction. CLIP \cite{radford2021learningtransferablevisualmodels} scale the training on weakly supervised (web-sourced image-text pair) data with cross-modal contrastive learning. Segment Anything \cite{kirillov2023segment} utilizes models and manpower together to increase the data scale and quality of segmentation data iteratively. 
% RL/Robotics algorithms \cite{embodimentcollaboration2024openxembodimentroboticlearning} use simulated/real environments to sample trajectory data.
These different types of data have various quality and cost. 

\begin{figure*}[htp]
\centering
\includegraphics[width=0.7\linewidth]{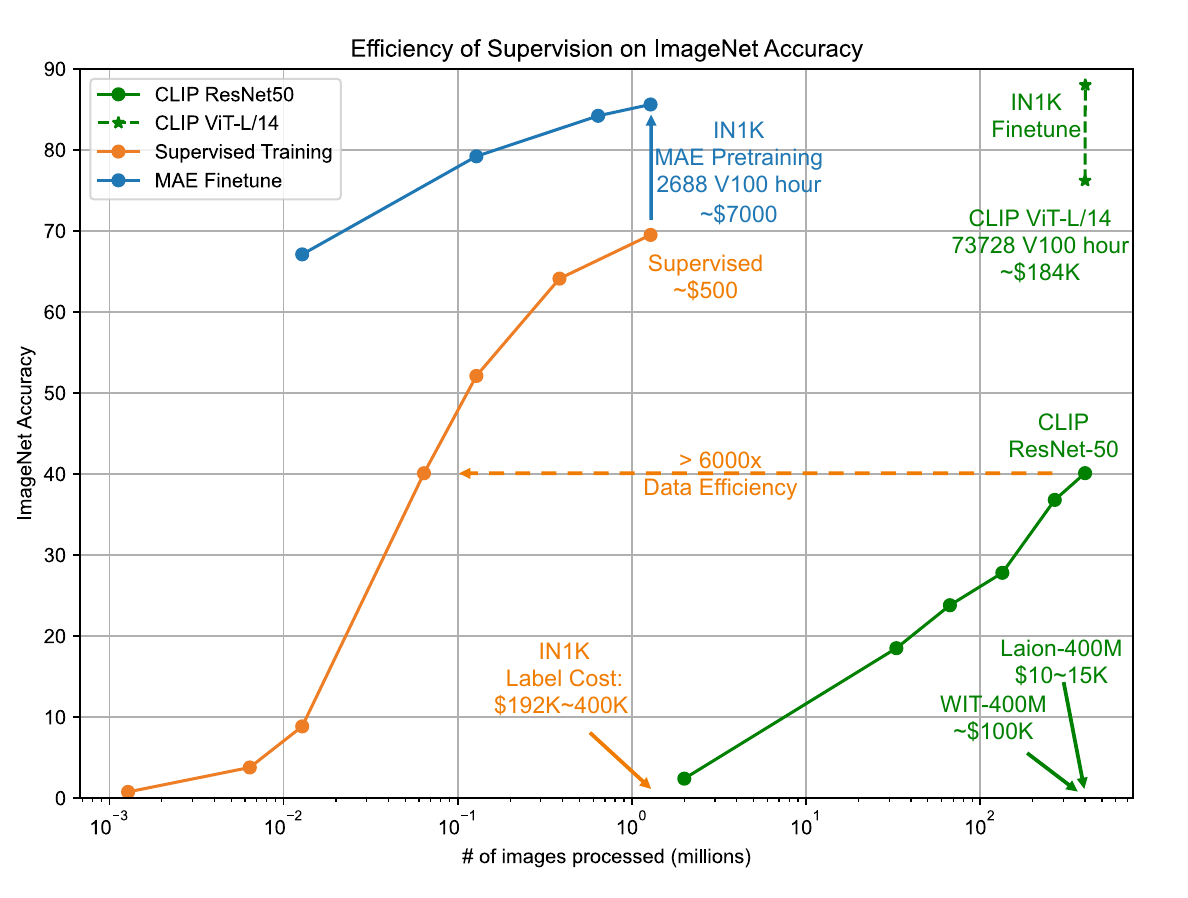}
\caption{The data scaling curve of different types of data. Supervised data has higher data efficiency than weak supervised data on specific downstream tasks (e.g. ImageNet), while incurring a higher collection and annotation cost.} 
\label{fig:data_scaling}
% \vspace{-10pt}
\end{figure*}

% 研究者发现的问题和改进后的当前方案：
% 从对下游性能提升的数据效率来看，有监督>弱监督>自监督，数据规模则是 有监督<<弱监督<<自监督, 三种范式是数据成本和训练成本的tradeoff。存在的问题:预训练（无监督数据）的scaling对于大部分人负担不起；弱监督可以但scaling的训练开销依旧较高；有监督和类似SAM-1B的范式依旧很贵，数据成本很高。
% BLIP, 数据质量提升能改进效率和性能；LLaMA系列对于GPT系列和早期系列也是如此。。。基于模型，模型加人工的数据质量提升以及数据量scaling都取得了对应效果，而提升数据质量在scaling中相当于提升监督程度。
As shown in Fig.\ref{fig:data_scaling}, regarding the task-specific/downstream data efficiency(performance against data amount) and data construction cost, supervised data $>$ weakly supervised data $>$ unsupervised/self-supervised data. The data scale and training cost are in a reversed order, where supervised $<<$ weakly supervised $<$ unsupervised/self-supervised. The three paradigms provide different tradeoffs between data collection/annotation and training costs. In general, a higher degree of supervision provides better data efficiency for downstream tasks with higher data cost per unit. Previously works like BLIP\cite{li2022blip} and Segment Anything \cite{kirillov2023segment} tried to boost the data quality with human+model annotation, which is somehow equivalent to improving the degree of supervision.

% 我们提出的方法和使用场景：网上以及下游实际在线场景中，数据其实在不断增长。无监督scaling的数据量级太大，训练带来的边际收益缩减较快；弱监督。。。；有监督标注成本。。。为了有效率的改进性能和同时低负担地维护高质量数据，我们设计了Info-Coevolution方法来在线地结合现有模型提升标注效率和数据扩增效率并高效迭代。。。并且提供了一种基于公共数据来增强下游数据的方法和对应的使用方式。Info-Coevolution可以同时用来应对高效率的在线数据质量提升和数据量扩展，并在过程中迭代模型形成高效的良性循环。

In online and downstream real-world scenarios, data continuously grows over time. The scale of unsupervised data could be immense, but the marginal gains from training diminish rapidly. Boosting its scale is not cost-efficient. Weakly supervised data involves a smaller overall volume with higher training efficiency per unit of data. In contrast, supervised data generally has the smallest total volume but incurs the highest annotation costs, leading to the most significant performance improvements for the same amount of data. As data grows in online scenarios, and usually the annotation cost is higher than training, it would be beneficial to improve the annotation efficiency and exlore the marginal benefits of supervised data.

% 之前方法问题；active cycle里有训练不现实，model based可能会fail；coreset没有利用模型的信息来做增量标注的筛选，所以不是最优
To boost the annotation/sample efficiency, active learning was proposed, which aims to select samples that better benefit training to annotate. However, many of the current active learning methods\cite{li2024surveydeepactivelearning} involve frequent model training with rounds of annotation, which makes the pipeline too complex for real-world applications. What's more, many active learning methods rely solely on the model's prediction for the sample's usefulness. It is prone to sample distribution problems on harder tasks. Their design of re-training + re-inference limits its application to large-scale data. Due to these limitations, active learning is not yet an efficient and robust enough solution for real-world scenarios. 
Coreset selection methods \cite{guo2022deepcorecomprehensivelibrarycoreset}, on the other hand, skip the model update and leverage the sample distribution to select samples used for training. Therefore, it fails to leverage the model-specific information 
unless a model trained on fully annotated data is obtained initially.
%unless you first get the already trained model (on fully annotated data).

To address these challenges, we propose Info-Coevolution, a novel and efficient online framework for selective data collection/annotation that integrates model-specific estimation with distribution awareness. Info-Coevolution enhances data annotation efficiency with minimal computational overhead. Leveraging Bayesian principles and our analysis of data locality, we propose a data-based information gain estimation strategy and Bayesian Prediction Fusion. This approach improves the model-specific sample selection while reducing the need for frequent model updates during the selection process.
The framework begins with an unsupervised pretrained backbone and a small, randomly initialized dataset, mimicking real-world scenarios. By utilizing an online Approximate Nearest Neighbor (ANN) structure, such as HNSW \cite{malkov2018efficientrobustapproximatenearest}, Info-Coevolution achieves logarithmic scaling for information gain estimation, enabling it to efficiently handle growing data and increasing selective annotation.

On ImageNet-1K, Info-Coevolution achieves lossless performance with only 68\% of the annotation cost, through just 4 rounds of continual supervised training (sum to $<$100\% full training cost; training from scratch gets the same lossless result). The computational overhead primarily arises from model inference after each training round, totaling approximately 10 A100 GPU hours for the entire process. Additionally, we developed an efficient batch sampling mechanism, allowing the selection process to be completed within 1 minute on datasets at a million-scale.
Furthermore, Info-Coevolution provides both qualitative and quantitative estimates of a data sample’s information gain, conditioned on the model and the previously collected data. This feature enables automatic stopping when performance gains plateau, eliminating the need for additional sample annotations to verify saturation.

%% file: sections/2-background.tex
\section{Related Works}
% datasets
% \textbf{Datasets}. (\textit{Introduce the famous dataset and their construction way and estimated cost here.}) Supervised datasets: higher building cost, high data efficiency, low extendability (cannot easily scale). Weakly-supervised/unsupervised datasets: lower building cost, medium data efficiency, high availability.

% compared to corset methods?
\textbf{Coreset Selection Methods} 
%\textit{Introduce coreset selection methods. Our main difference: is online extendable, coreset is offline; we do not need the full data to then do the selection. More related to Dataset Growth \cite{qin2024datasetgrowth}}
Coreset selection methods focus on filtering out low-quality or redundant samples, while reserving the most representative ones in the target dataset.
The core of these approaches lies in the elaborate selection criteria, including geometry-based~\cite{sener2017active,sinha2020small}
, error-based~\cite{toneva2018empirical,paul2021deep}
, decision-boundary-based~\cite{ducoffe2018adversarial,margatina2021active}
, uncertainty-based~\cite{coleman2019selection}
, gradient-matching~\cite{mirzasoleiman2020coresets,killamsetty2021grad}, bilevel optimization~\cite{killamsetty2021glister}
and submodularity-based methods~\cite{iyer2021submodular,zhou2023datasetquantization}.
Some of the approaches~\cite{sener2018activelearningconvolutionalneural} also combined the strength of active learning to make further improvements.

\begin{figure}[h]
\centering
\includegraphics[width=1.0\columnwidth]{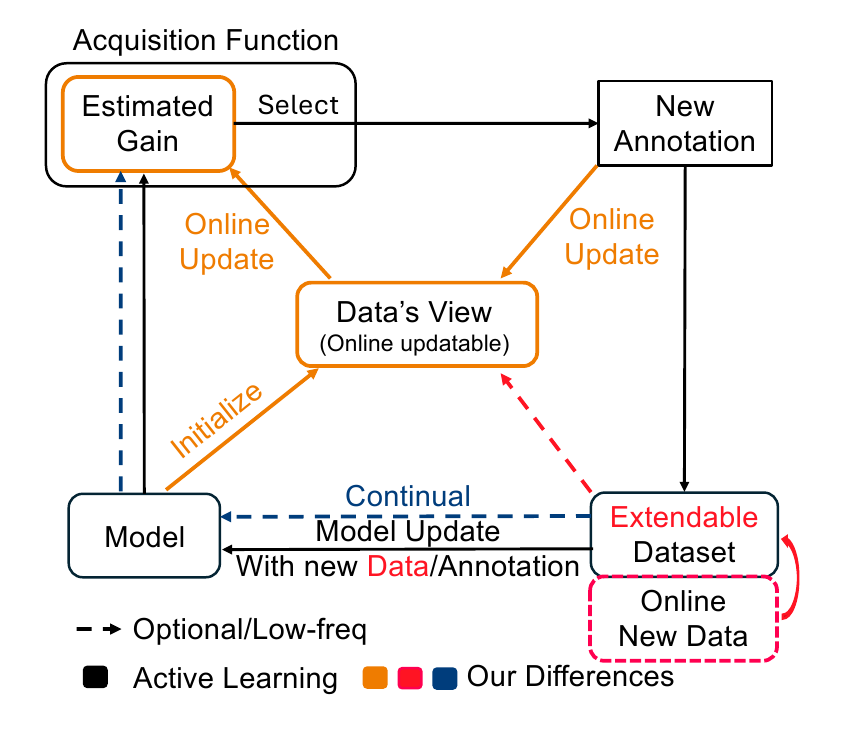}
\vspace{-20pt}
\caption{A comparison of our method's difference to active learning. Generally, active learning loops over 1. selecting samples, 2. annotating, 3. updating the model, and selecting samples again (with the updated model). In contrast, our method doesn't have to update the model frequently. We can update the model optionally with continual supervised learning at a much lower frequency (so the training cost is also low). And our algorithm is automatically class-balanced due to distribution awareness, therefore it doesn't suffer distribution shift as active learning.} 
\label{fig:active_compare}
\end{figure}

\textbf{Active Learning and Semi-Supervised Learning} Active learning \cite{hide2020activesurvey,smith2023predictionorientedbayesianactivelearning,li2024surveydeepactivelearning} and semi-supervised learning \cite{sohn2020fixmatchsimplifyingsemisupervisedlearning,zhang2022flexmatchboostingsemisupervisedlearning,wang2023freematchselfadaptivethresholdingsemisupervised,cai2022semisupervisedvisiontransformersscale} are two complementary strategies for reducing the reliance on large-scale supervised datasets in machine learning. Active learning focuses on iteratively selecting the most informative samples for annotation, enabling models to achieve higher performance with fewer labeled examples by optimizing the data-labeling process. On the other hand, semi-supervised learning leverages a large pool of unlabeled data alongside a smaller labeled subset, using techniques such as pseudo-labeling, consistency regularization, or generative models to propagate label information and improve generalization. Together, these paradigms aim to maximize model efficiency and performance in data-scarce scenarios, often bridging the gap between fully supervised learning and real-world constraints.
Segmenta anything, in a way similar to active learning, conducts full annotation each round with different quality and re-train the model for each round on increased amount/portion of data. Different to these works, our method targets an efficient online selective data collection/annotation process, with far less cost in the loop. An illustration of the difference is shown in Fig.\ref{fig:active_compare}.

% Previous work Dataset Growth

%% file: sections/3-method.tex
\section{From the Perspective of Information Gain}
In this section, we first derive from the theoretical aspect of how to estimate the information gain of a sample to a dataset. Previously, an exact information gain was mainly defined upon an attribute of a decision tree. Here we extend it to a more generalized form and provide a way to estimate it on broader tasks, e.g. a sample's information gain with respect to image classification (datasets). We analyze the influence of locality in the space to improve both the estimation quality and efficiency.
Then we formulate the selective data selection/annotation problem as a recursive online problem, each time adding one (batch of) sample/annotation to an already collected dataset. Based on this problem setup and our theoretical analysis, we propose our algorithm Info-Coevolution for efficient online selective annotation and dataset construction.

\begin{theorem}
\label{lemma_sim}
For neural networks that generate representation and then make the prediction (for model $M=g\circ f$), at certain distances $\epsilon$ from a sample point $x$ in the feature representation space,
the similarity of samples in feature space also leads to similar model predictions ($\forall \delta, \exists \epsilon \ s.t. \forall x,x', |f(x')-f(x)|\leq\epsilon \Rightarrow |g(f(x'))-g(f(x))|\leq \delta$). 
\end{theorem}
We show a proof of Theorem \ref{lemma_sim} with both L2 distance and cosine distance in Appendix. This theorem implies a kind of linearity among near-neighbor samples' predictions, and explains one of the reasons that KNN predictions work. We will leverage this theorem to design a parametrized estimation of a sample's information gain with locality, extendability, and efficiency.

\subsection{Information carried by sample to the dataset}
In predictions with a decision tree $T$, information gain $IG$ with condition $A$ is defined as:
\begin{equation}
    IG(T,A) = H(T)-H(T|A),
\end{equation}
which is the reduced entropy given condition A. Here, we extend this information gain to broader tasks. Assuming the samples we are curious about (target distribution) is from distribution $\rho$, which is either uniform in the whole distribution space or IID to train data, then the information gain of adding a sample $z=\{z_x,z_y\}$ to a dataset $\mathcal{D}$ sampled from $\rho$ is:
% \begin{align}
%     IG(\mathcal{D},z)= H(\mathcal{D})-H(D+z) = \\ \int_{x,\|x-z\|\leq \epsilon }{(H(x)-H(x|z))\rho(x)dx},\\
% or\   H(\mathcal{D})-H(\mathcal{D}+z) = \\ \mathbb{E}\left[\frac{1}{|D|}\sum_{x,sim(x,z)\geq \delta}{(H(x)-H(x|z))dx} \right] \ if \ IID
% \end{align}
\begin{align}
    IG(\mathcal{D}, z) &= H(\mathcal{D}) - H(\mathcal{D} + z) \notag \\
    &= \int_{x, \|x - z\| \leq \epsilon} \big(H(x) - H(x | z)\big) \rho(x) \, dx, \\
    \text{or } \, IG(\mathcal{D}, z) &= H(\mathcal{D}) - H(\mathcal{D} + z) \notag \\ \textit{(if IID)} \quad
    &= \mathbb{E}\left[\frac{1}{|\mathcal{D}|} \sum_{x, \operatorname{sim}(x, z) \geq \delta} \big(H(x) - H(x | z)\big) \right].
\end{align}
Where $\epsilon \in \mathbb{R}$ is the distance threshold we decide to use, and $\delta \in [0,1]$ is a similarity threshold. This distance threshold is inherited from Theorem\ref{lemma_sim}, where two sample's prediction has a low correlation beyond a certain distance. During estimation, the $H$ here is estimated on parameterized $M_\theta$, and the $||\cdot||$ is defined within space of $f$.

Then a question is how to estimate $H(x|z)$ given $x\neq z$. As assumed, in certain distances the linearity dominates so that we can do interpolation and linear combination on logits/predictions. Within this space where linearity holds, interpolation between $z$ and $x$ where $y_z=y_x$ also gives the same label prediction, but with an increased certainty. Interpolation between $z$ and $x$ where $z_y\neq x_y$ will split the prediction over the two, which needs careful calculation. However, the interpolation itself is an estimation based on linearity assumption. As different samples' predictions could have different certainty/confidence, we should take the confidence (probability of prediction being true) into consideration of the interpolation.

% model's view (cleaner prediction), dataset's view (knn), they provide independent and different view with different confidences.

\subsection{Confidence of label}
By defining a confidence $\alpha$ for the prediction of a sample, we can integrate this value into the interpolation process as a weighted NN prediction:
\begin{equation}
\label{eq:knn_pred}
    \mathbf{y_{estimation}(z)} = \frac{\sum_{x,sim(x,z)\geq \delta}\alpha_x Sim(x,z)\mathbf{y_x}}{\sum_{x,sim(x,z)\geq \delta}\alpha_x Sim(x,z)}
\end{equation}
where $\mathbf{y_x}$ is the probability prediction vector of x's annotation. Then the entropy of a point z from data's view can be calculated based on it. Moreover, the semi-supervised learning's label and unlabeled datasets can be transferred into a continuous space defined on confidence $\alpha \in [0,1]$, and we can estimate the confidence of an annotator and use models to perform annotation. An easy mapping in probability space can directly map accuracy to confidence:
$\alpha = (acc-acc_{random})/(1-acc_{random})$. One primary goal of semi-supervised learning under this interpretation could be making the $\alpha$ value converge while training the model with corresponding confidence without collapse.

\begin{figure*}[htp]
\centering
\includegraphics[width=1.0\linewidth]{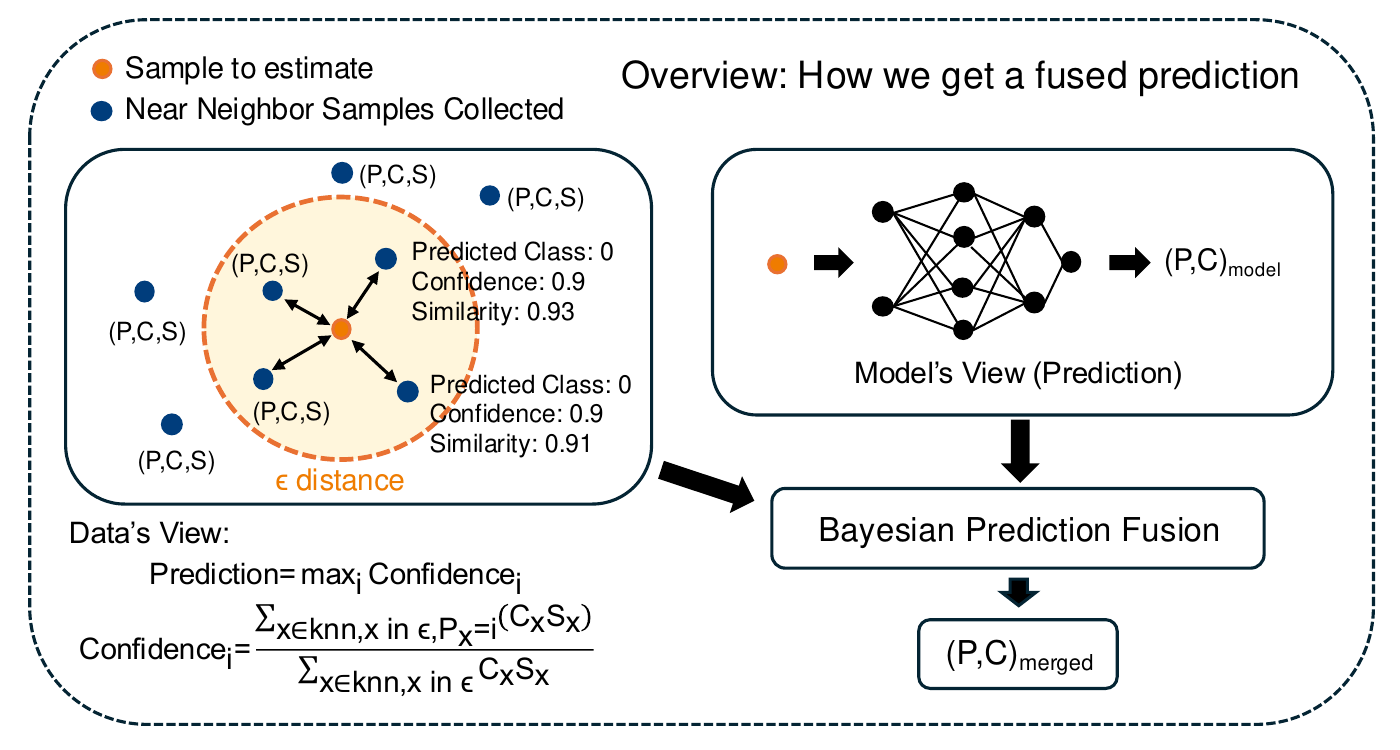}
\vspace{-20pt}
\caption{The overview of how we estimate a sample's confidence with Bayesian Fused prediction from both model's view and data's view. Generally, this estimation gives a better evaluation of a sample's uncertainty/confidence than using model prediction only.} 
\label{fig:bayesian_fused}
\end{figure*}

\subsection{Bayesian prediction fusion}
Therefore, one key question in semi-supervised learning is interpreted as $\alpha$ value convergence. Suppose we have two independent predictors with confidence $\alpha_1$ and $\alpha_2$, when the two predictors have same prediction $y$ for sample $x$, the confidence value $\alpha_{y}$ of sample $x$ should be annotated with $y$ become
\begin{align}
\label{eq:converge}
    \alpha_{y} &= P(\frac{Predictions\ are\ true}{Predictions\ match}) \notag \\
    &= \frac{\alpha_1*\alpha_2}{\alpha_1*\alpha_2+P(match\ with\ other\ label)}
\end{align}
In classification tasks, $P (match\ with\ other\ label) \in [0,(1-\alpha_1)(1-\alpha_2)]$ depending on distribution over classes. In the binary classification case $P (match\ with\ other) = (1-\alpha_1)(1-\alpha_2)$, it requires smaller prediction confidence is larger than 0.5 so that the sample confidence is increased: $min(\alpha_1,\alpha_2) > 0.5 \Rightarrow \alpha_{xy} > max(\alpha_1,\alpha_2)$; if assuming the matching distribution is uniform over $c$ classes, then $P (match\ with\ other) = (1-\alpha_1)(1-\alpha_2)/(c-1)$ and $min(\alpha_1,\alpha_2) > 1/c$ is the condition. If no matching prob distribution is in other classes, then its value is 0. To sum up, using $(1-\alpha_1)(1-\alpha_2)$ gives a lower bound estimation on $\alpha_{xy}$.
These conditions can be utilized to make the alpha value converge, or re-estimate $\alpha$ value for samples in a dataset.

When predictions from different views diverge, if one is confident above the threshold and the other is not, use the max one for the highest confidence. If both predictions are confident, then the sample's confidence is reduced. Supposing $\alpha_1 > \alpha_2$, then
\begin{align}
\label{eq:diverge}
    P(p_1)&=\frac{P(p_1\ is\ true)}{P(predictions\ unmatch)} \notag \\
    &=\frac{\alpha_1(1-\alpha_2)}{\alpha_1(1-\alpha_2)+P(other)}
\end{align}
where $P(other)\in [0,(1-\alpha_1)\alpha_2]$. In the case $p_2$ takes all the remaining confidences from the first predictor, $P(other)=(1-\alpha_1)\alpha_2$; if assuming the matching distribution is uniform over c classes, $P(other)=(1-\alpha_1)\alpha_2/(c-1)$. If there are no prob distribution unmatching classes on other classes, $P(other)=0$. Similarly, using $P(other)=(1-\alpha_1)\alpha_2$ gives a lower bound on the updated confidence. The prediction divergence from confident predictions means $max(p_1)\neq max(p_2) \Rightarrow H(p')>max(H(p_1),H(p_2))$. Reducing the corresponding confidence can help with re-convergence.

% The $\tau$ value is usually used in semi-supervised learning to encourage the model to progressively learn and label with confident guesses. However, previously there was no explicit way to reduce the labeled set. Label and Unlabel are discrete \{0,1\} confidence for learning, and once a label is wrongly predicted it may never be flipped again. Therefore we propose to introduce a sample-wise confidence instead of using just model-prediction, which is an ensembled result from both model and knn predictions and allows for reducing sample confidence as described above.

% \subsubsection{Annotation cleaning as part of confidence convergence}
% Human annotation confidence varies across tasks. On classification, it is about 0.9$\sim$0.95, and its value is reflected as soft labeling. By making this value trainable when the model is stable or evaluating it on the cleaned subset, we will be able to find sample annotation confidence and noisy annotations. (while each time the noisy samples found may differ)

\subsection{Benefit of reannotating one sample}
As the model is to absorb the data distribution, the dataset itself also carries a prediction based on the near neighbor prediction within the distance where linearity holds. And sample's confidence changes during turns of iteration. Therefore, taking the derivation from previous sections, when re-annotating a sample, the expected gain is 
\begin{align}
\label{infomation_gain_calculation}
    \mathbb{E}[IG(annotate\ z)] =max\big(min(H(\alpha_z),& \notag \\
      H(knn\ predictions)) - H(\alpha_{ann}),0\big)&
\end{align}

If using entropy, one can use $H(\alpha)=-\alpha log(\alpha) - \sum_{c-1}\frac{1-\alpha}{c-1}log\frac{1-\alpha}{c-1} \simeq -\alpha log(\alpha) - (1-\alpha)log(1-\alpha)$ as an estimation neglecting the constant term. In practice, we found using $H(\alpha)=\alpha$ is a feasible proxy to choose samples with less computation and higher compatibility. For sample selection, we only need to know the positivity and relative magnitude of the samples' information gain, therefore the confidence itself is a good proxy.

%How to measure on other loss?

% \subsection{When dealing with group of samples}
% For single-sample addition/re-annotation, the information gain is easy to calculate and optimize. However, when there is a group of samples to add/re-annotate, the corresponding gain is harder to decide. An optimal (greedy) choice for one step will not lead to an optimal global solution.

% Fortunately, we can formulate the choice over the set problem as a Discrete Quadratic Programming problem. We can have
% \begin{equation}
%    maximize -\frac{1}{2}x^TQx+c^Tx, s.t. \ Ax = a,
% \end{equation}
% where in our setting, x is $\{0,1\}^n$ representing the selection of samples, Q is the similarity of samples, and c is the information gain of each sample, A is all 1, a is the number of samples selected. An optimal solution to this equation would take exponential time. We take this calculation within localized space where linearity dominates so that the similarities directly penalize local information predictions, the information gain is with respect to model prediction or no-prior belief, and the time cost is more tractable. Based on this formulation, we can add/re-annotate samples with near-optimal choices maximizing information gain for current knowledge (dataset and model). 

\subsection{The Algorithm}
\label{sec:algorithm}
% (TO BE replaced by formal algorithm later) \\

\textbf{Selective Annotation} When data to be annotated is greater than the capacity, for a sample $z=(x_i)$, to estimate the gain of annotating it given already collected data D, we have model $m$ and get the feature $f(z)$, model prediction $y_m$, confidence $c_m=(p_m-1/num\_class)/(1-1/num\_class)$. We then retrieve k nearest neighbors of z (with f and ANN search) and use Eq.\ref{eq:knn_pred} to get a (knn) data-based prediction $y_{knn},c_{knn}$. Then we merge two predictions from model and dataset with Eq.\ref{eq:converge} and Eq.\ref{eq:diverge} and get $y_{merged},c_{merged}$. The annotation gain of sample z is then calculated as $c_{ann}-c_{merged}$ where $c_{ann}$ is the average accuracy of the annotator. For batched selection, one can sample with probability proportional to the gain and drop redundant samples.

\begin{figure}[h]
\centering
\includegraphics[width=1.0\columnwidth]{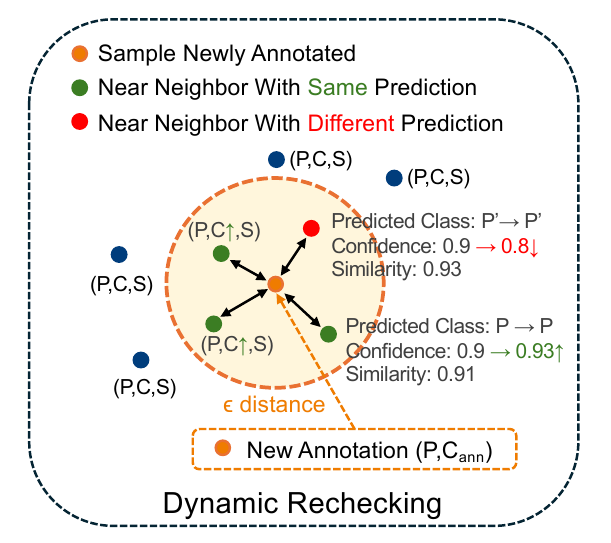}
\vspace{-20pt}
\caption{The idea of dynamic rechecking. When we get new annotation(s), we can update the sample estimation within the $\epsilon$-distance to better reflect the gain. It will automatically balance classes and sample density. This step is efficient with klog(n) time.} 
\label{fig:dynamic_rechecking}
\end{figure}

\textbf{Dynamic Rechecking} After getting a new annotation $y_i$ for $z=(x_i)$ with confidence $c_{ann}$, we retrieve k nearest neighbors of z. For neighbors within the distance threshold, we recalculate their data's view prediction and update their (P,C) with our Bayesian Prediction Fusion Eq.\ref{eq:converge} and Eq.\ref{eq:diverge} and update their gain. Each time we select the (batch) sample with the highest annotation gain to annotate. This design is to avoid redundancy and ensure class balance during selective annotation, as near samples with similar predictions will be updated with a lower gain, while near samples with different predictions will be updated with a higher gain. We use online-ANN based near neighbour search so that this process is efficient and extendable.

% \begin{algorithm}[h]
%    \caption{Placeholder}
%    \label{alg:online_sample_estimation}
% \begin{algorithmic}
%    \STATE {\bfseries Input:} data $(x_i,y_i,c_i)$, model $m$, HNSW index $h$, 
%    \REPEAT
%    \STATE $f,y',c=m(x_i)$ \COMMENT{Extract feature, prediction and confidence}
%    \STATE Lookup Near Neighbour $NN$ of $f$ in $h$
%    \STATE 
%    % \FOR{$i=1$ {\bfseries to} $m-1$}
%    % \IF{$x_i > x_{i+1}$}
%    % \STATE Swap $x_i$ and $x_{i+1}$
%    % \STATE $noChange = false$
%    % \ENDIF
%    % \ENDFOR
%    \UNTIL{$noChange$ is $true$}
% \end{algorithmic}
% \end{algorithm}

\subsection{Using Public Data to Enhance Downstream Task}
\label{sec:dataset_enhance}
To further benefit downstream tasks and evaluate our method in the data/annotation growth setting, we construct a superset from multiple open-sourced (and mostly web-sourced) datasets (CC3M\cite{sharma2018conceptual}, CC12M\cite{changpinyo2021conceptual12mpushingwebscale}, SBU\cite{NIPS2011_5dd9db5e}, COCO\cite{chen2015microsoftcococaptionsdata}, Visual Genome\cite{krishna2016visualgenomeconnectinglanguage}, Laion-400m\cite{schuhmann2021laion400mopendatasetclipfiltered}) and retrieve related samples to enhance downstream tasks. To reduce the peak GPU/RAM memory, we subsample 1\% samples from all the samples and retrieve their feature using BLIP\cite{li2022blip}. Then we use K-means to get 500 clusters. We then follow \cite{qin2024datasetgrowth} to do ANN-based de-redundancy and filtering within each cluster and construct the corresponding ANN index for later retrieval. This allows a higher parallelism of the algorithms.
For future data scale extension, either distributed HNSW or hierarchical recursive clustering plus cluster-wise HNSW is a feasible solution. 

On ImageNet-1K, we use image embedding from ImageNet-1K samples to retrieve the k nearest neighbor in the superset, and then de-redundant the samples while filtering retrieved samples with cosine distance larger than 0.2.

% \subsection{Human-model Cooperated Annotation}
% With data from \ref{sec:dataset_enhance}, there are actually two ways to retrieve the related data. One way is to use the image embedding; another way is to use the image captions to turn them into downstream supervision signal. Combining with the algorithm in \ref{sec:algorithm}, we can boost the data quality of these weak supervised samples to a degree closer to human annotation.

%% file: sections/4-analysis.tex
\section{Experiments}
% 几个部分：imagenet主实验的曲线（有监督的，我们的，和半监督结合的，半监督的，超越半监督的）；
% Further scale的
% 我们方法不依赖于任何强分类模型，不需要提前标注好的数据来进行类别平衡（相比于dq）
\subsection{Datasets and Implementation Details}
We evaluate our method on ImageNet-1K \cite{imagenet5206848}, CIFAR10/100\cite{cifar10,cifar100}, StanfordCars, Food-101\cite{Bossard2014fool101}, SVHN\cite{yuval2011svhn} under different annotation ratios and settings. We further extend the training data with data from CC3M, CC12M, SBU, Visual Genome, COCO, and LAION-400m to study the effect of scaling unlabeled data.
%TODO: CITATIONs here

\textbf{Implementation Details.} The experiments on ImageNet-1K follows Semi-ViT \cite{cai2022semisupervisedvisiontransformersscale}, using ViT \cite{dosovitskiy2021imageworth16x16words} model with MAE \cite{he2021maskedautoencodersscalablevision} pretrained backbone to conduct supervised and semi-supervised training. All other details can be found in the Appendix.

\subsection{Data/Annotation Scaling}

% \begin{table}[h]
%     \centering
%     \caption{Training ViT-Large on ImageNet-1k with different annotation ratio. Info-Coevolution gets a 1.3\% accuracy improvement in 10\% ImageNet annotation setting with the proposed prioritized selective annotation.}
%      \vspace{10pt}
%     \begin{tabular}{cc|ccc}
%     \toprule
%     Method & Setting & 1\% & 10\% & 50\% \\
%     \midrule
%      Random & Supervised & 67.1 & 79.2 & 84.2\\
%      Ours & Supervised & - & \textbf{80.5} & \textbf{85.1} \\
%      \midrule
%     Random & Semi & 77.3 & 83.3 & 84.8\\
%      Ours & Semi & -  & \textbf{83.7} & \textbf{85.3}\\
%          \bottomrule
%     \end{tabular}
    
%     \label{tab:ImageNet}
% \end{table}

\begin{figure}
    \centering
    \includegraphics[width=1.0\columnwidth]{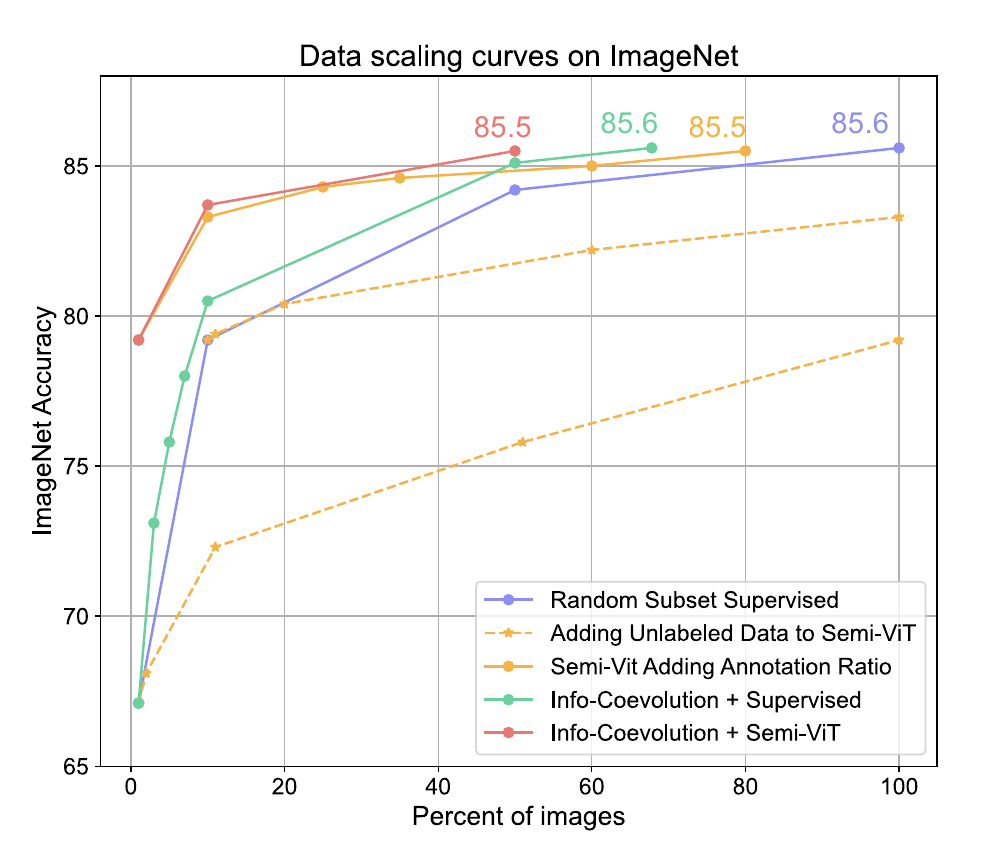}
    \vspace{-20pt}
    \caption{The scaling curve of different schemes. Info-Coevolution improves data efficiency for both Supervised/Semi-supervised settings, and gets lossless performance at 68\% and 50\% annotation ratios respectively.}
    \label{fig:scaling}
    \vspace{-10pt}
\end{figure}

\textbf{ImageNet-1k Results} We show our improvement in data/annotation efficiency here in Fig.\ref{fig:scaling} and Tab.\ref{tab:continual_ann}. On 10\% data setting of ImageNet, our selective annotation can increase the accuracy by 1.3\% compared to the random baseline. With \textbf{only 68\%} annotated samples from ImageNet-1k, we can achieve \textbf{lossless} performance (85.6\% Acc), surpassing the 85.5\% Acc of Semi-ViT with 80\% labeled data and 20\% unlabeled data. What's more, we have an \textbf{automatic stop criterion} when the marginal gain of annotating more samples is negligible. The 68\% ratio is given by the algorithm itself when it suggests to stop selecting samples, instead of a annotation ratio tuning. It can be seen that Info-Coevolution is compatible with Semi-Supervised learning, where Semi-ViT trained with 50\% ImageNet-1K annotations selected by Info-Coevolution can achieve an almost lossless result (85.5\%).
Moreover, Info-Coevolution is compatible with continual supervised training, without introducing a distribution shift. 

% 加一下和DQ的比较并且强调我们方法不用提前得到所有数据和标注。

\begin{table}[h]
    \centering
    \caption{Accuracy of training ViT-Large on ImageNet-1k with continually increasing annotation and doing continual training. Info-Coevolution gets a 1.3\% accuracy improvement in 10\% ImageNet supervised setting with the proposed prioritized selective annotation.}
    
    \resizebox{\columnwidth}{!}{
    \begin{tabular}{cc|ccccc}
    \toprule
    Method & Setting & 1\% & 3\% & 5\% & 7\% & 10\% \\
    \midrule
     Random & Supervised & 67.1 & 72.5 & 74.6 & 76.6 & 79.2\\
     Ours & Supervised & - & 73.1 & 75.8 & 78.0 & 80.5 \\
     \bottomrule
    \end{tabular}
    }
    \label{tab:continual_ann}
\end{table}

\begin{figure}[h]
    \centering
    \includegraphics[width=1.0\linewidth]{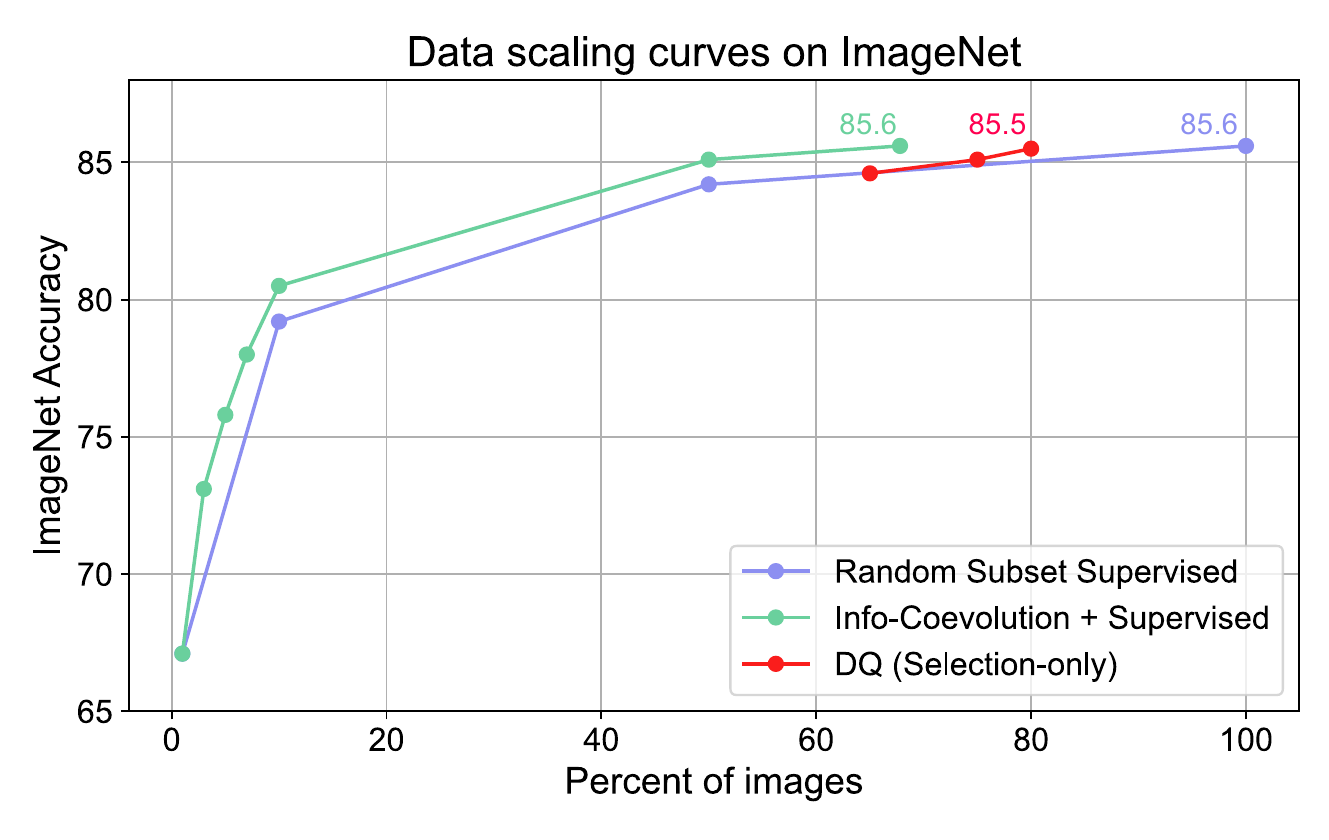}
    \vspace{-20pt}
    \caption{Compare with corset-selection SOTA method Dataset Quantization on IN1K.}
    \label{fig:dq_compare}
\end{figure}

\textbf{Comparison with Coreset Selection} In Fig.\ref{fig:dq_compare}, we compare with previous SOTA method Dataset Quantization\cite{zhou2023datasetquantization}. With selection-only (not adding the MAE reconstruction), DQ achieves 85.5\% Acc with 80\% labeled data. It shows that beyond the extendability of Info-Coevolution, it can also serves as a good coreset selection method.

\begin{table}[h]
    \centering
    \caption{Scaling the dataset with retrieved data from our constructed superset and training with Semi-ViT. Extending the unlabeled data can improve performance while using our selection would improve the data efficiency of unlabeled data.}
    \begin{tabular}{cc|c}
       Labeled Data  & Additional Data & Acc \\
    \toprule
        1.28M & 0 & 85.6 \\
        1.28M & Unlabeled 2M & 86.0 \\
        1.28M & Our selected unlabeled \textbf{1M} & 86.0 \\
    \bottomrule
    \end{tabular}
    \label{tab:data_scaling}
\end{table}

Tab.\ref{tab:data_scaling} illustrates the effect of data scaling for further scaling semi-supervised training with unlabeled data. With additional unlabeled data, the acc can be further increased by 0.4\% with semi-supervised training. For unlabeled data selection, we adjust the equation to better capture useful unlabeled data. See Appendix for detail. It use half of the unlabeled data to enhance the performance as using all unlabeled data.

\textbf{Generalization and Robustness} Theoretically, the effectiveness of Info-Coevolution is model-agnostic and dataset-agnostic. We here verify its generalization and robustness across different datasets, architectures and other semi-supervised framework.

We present our experimental result on CIFAR10 semi-supervised training with Fixmatch\cite{sohn2020fixmatchsimplifyingsemisupervisedlearning} and ResNet-50x4 in table \ref{tab:cifar_fixmatch}. Info-Coevolution improves the data efficiency and further reduces the annotation amount to achieve lossless performance (95.85\% acc) from 4000 to 1000.

\begin{table}[h]
    \centering
    \caption{ResNet-50x4 with Fixmatch on CIFAR10}
    \begin{tabular}{c|cccc}
    \toprule
    Data Selection & 250 & 1000 & 4000 & Full\\
    \midrule
    Random  & 94.95 & 95.59 & 95.85 & 95.85 \\
    Info-Evolution  & 95.39 &\textbf{ 95.85} & - & - \\
    \bottomrule
    \end{tabular}
    \label{tab:cifar_fixmatch}
\end{table}

\begin{figure*}[htp]
\centering
\includegraphics[width=1.0\linewidth]{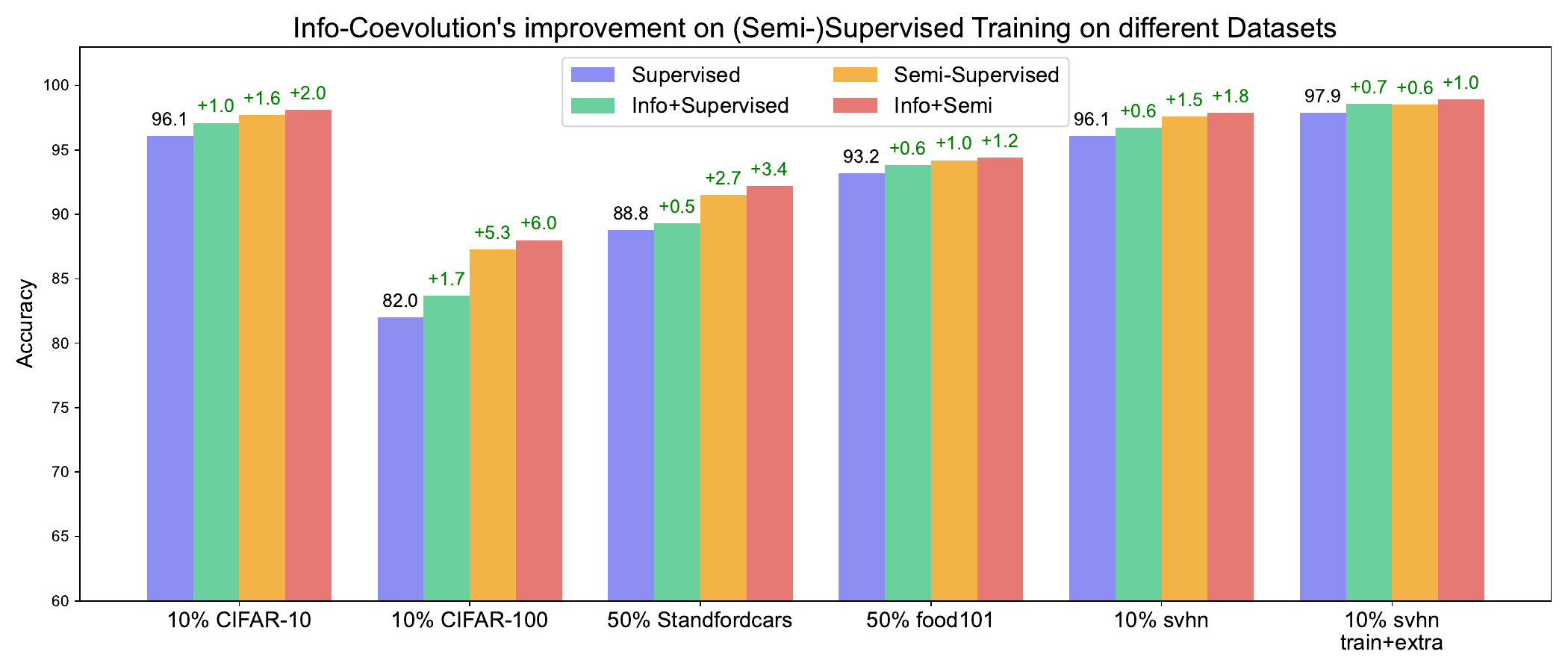}
\vspace{-20pt}
\caption{Info-Coevolution consistently improve the annotation efficiency across different datasets with both supervised/semi-supervised setting.} 
\label{fig:data_generalization}
\end{figure*}

To analyze the generalization of Info-Coevolution across different datasets, we use Info-Coevolution with both supervised training and semi-supervised training by adapting pretrained ViT-L on CIFAR-10, CIFAR-100, Standfordcars, fool101 and SVHN. As shown in Fig.\ref{fig:data_generalization}, Info-Coevolution consistently improved both supervised and semi-supervised training performance on all these datasets.

% \begin{table}[h]
%     \centering
%     \caption{Accuracy of training ViT-Large on other datasets with different annotation ratio. Info-Coevolution gets a 1.3\% accuracy improvement in 10\% ImageNet supervised setting with the proposed prioritized selective annotation.}
%     \resizebox{\columnwidth}{!}{
%     \begin{tabular}{ccccc}
%     \toprule
%     Dataset & Data Selection & Supervision & Sample & Acc(\%) \\
%     \midrule
%     CIFAR10 & Random & Supervised & 1000 & 96.5\% \\
%     CIFAR10 & Info-Coevolution & Supervised & 1000 & 97.1\% \\
%     CIFAR10 & Random & Semi-supervised & 1000 & 96.5\% \\
%     CIFAR10 & Info-Coevolution & Semi-supervised & 1000 & 97.1\% \\
%     \bottomrule
%     \end{tabular}
%     }
%     \label{tab:other_dataset}
% \end{table}

\subsection{Ablation Experiments}

\input{tables/ablation_components}

\textbf{Ablation of Components.} As our algorithm involves fusing the prediction of model and KNN predictions with dynamic rechecking, we here ablate their corresponding influence on performance in Tab.\ref{tab:abl_comp}. The experiment is to choose annotation for 10\% ImageNet data with 1\% random data trained model. It can be seen that using only model prediction for sample gain estimation could fail to beat the random baseline (78.5\% compared to 79.2\%). As analyzed, purely model-uncertainty-based sample selection is unaware of distribution, which could lead to distribution bias in hard problems.
Using our data's view prediction gives a better prediction of information gain and gets 0.6\% performance improvement; the dynamic rechecking which introduced locality consideration effectively mitigate the problem of model-based sample selection, improving performance by +0.5\% (compared to model-only -0.7\%).
Our proposed multiview prediction fusion combines both model and KNN prediction to better decide the samples with higher entropy. Using the fused prediction alone will get a severe distribution balance problem, which lower down the performance by 2.1\%. With dynamic rechecking involved to balance both density and class balance, it gets a 1.0\% ACC improvement compared to the random baseline.

\begin{figure}[ht]
    \centering
    \includegraphics[width=1.0\columnwidth]{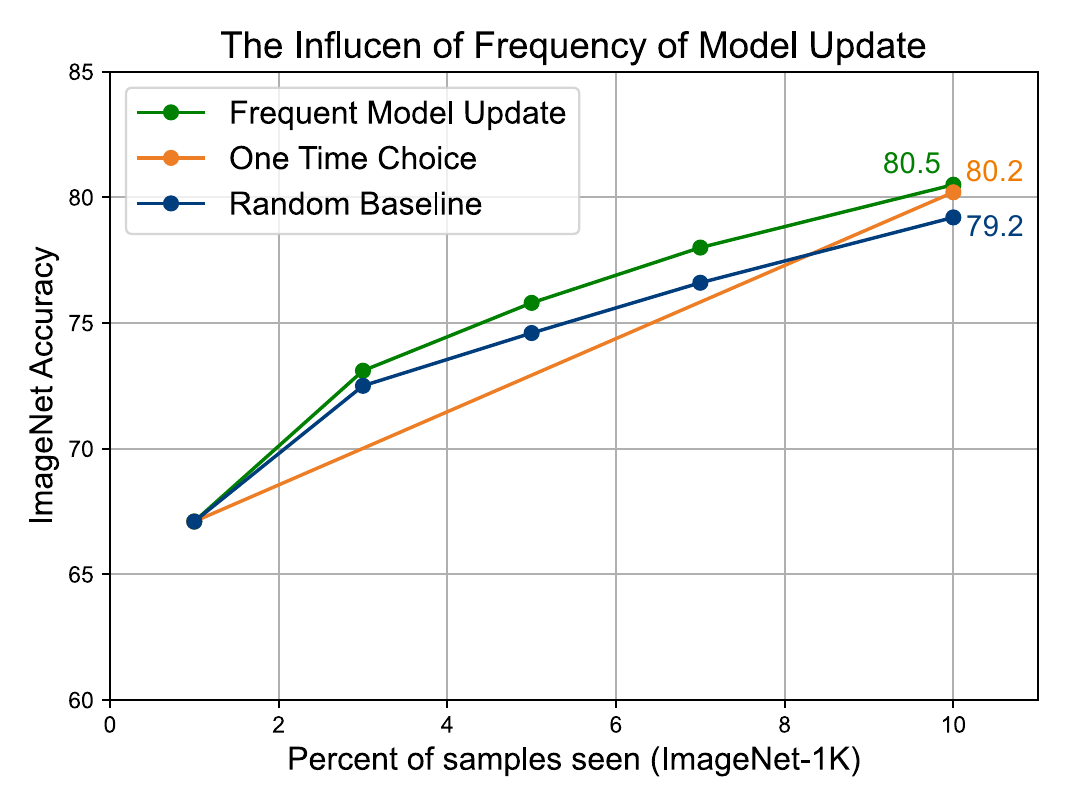}
    \vspace{-20pt}
    \caption{Ablation of model update frequency. When updating the model more frequently during selection, it shows improved data efficiency.}
    \label{fig:freq_abl}
    \vspace{-10pt}
\end{figure}

\textbf{Ablation of Model Updating Frequency.} We also study the influence of model fine-tuning frequency on annotation-efficiency. In Fig.\ref{fig:freq_abl}, we can see that, when we add additional model updates during the loop, Info-Coevolution can select a better annotation set at the annotation ratio. When directly selecting 10\% (0.128M) annotations for ImageNet-1k with model trained on 1\% data, the performance is 80.2\%; after adding two model updates in the middle, we get 80.5\%.

%% file: tables/ablation_components.tex
\newcommand{\red}[1]{$_{\color{Red}\downarrow #1}$}
\begin{table}[h]
    \centering
    \small
    \caption{Ablation of proposed components in the framework (1\% model to select 10\% annotation).}
    \vspace{5pt}
    % \resizebox{0.48\textwidth}{!}{
    \begin{tabular}{ccc|c}
        \toprule
        \multicolumn{3}{c|}{Component} & \multicolumn{1}{c}{ImageNet Acc.} \\
         Model & Data & Dynamic &  10\% label \\
         \midrule
         \checkmark & &   &   78.5\red{0.7} \\
          & \checkmark &   &  79.8\green{0.6} \\
          \checkmark & \checkmark &  & 77.1\red{2.1} \\
         \checkmark & & \checkmark  & 79.7\green{0.5}  \\
          & \checkmark & \checkmark  & 79.8\green{0.6}  \\
         \checkmark & \checkmark & \checkmark  & \textbf{80.2\green{1.0}} \\
         \midrule
         \multicolumn{3}{c|}{Random Select} & 79.2 \\
         \bottomrule
    \end{tabular}
    % }
\label{tab:abl_comp}
% \vspace{-20pt}
\end{table}

%% file: sections/5-conclusion.tex
\section{Conclusion}
In this work, we extend Information Gain estimation to classification tasks with locality consideration. We proposed a novel online efficient algorithm Info-Coevolution, to increase the annotation efficiency of supervised/semi-supervised training. Info-Coevolution is able to save the annotation ratio by 32\% on ImageNet with a lossless performance and is compatible with Semi-Supervised learning to achieve almost lossless performance with 50\% annotation. We also demonstrate how to enhance the downstream task dataset with open-source data. As an online method, Info-Coevolution is efficient and extendable for real-world applications.

\textbf{Limitations and future works} Our work mainly considers the same training data distribution as the target distribution on the classification task. Using the large-scale weakly supervised data may be using different training data distribution to target.
For unlabeled data and wealy-supervised data, it is possible to extend the framework further while we have only done a preliminary study. Tasks beyond classification are also worth further exploration.